\documentclass[runningheads]{llncs}
\usepackage{graphicx}
\usepackage{color}
\usepackage{xcolor}
\usepackage{textcomp}
\usepackage{amsmath}
\usepackage{amssymb}
\usepackage{amsfonts}
\usepackage{cite}
\usepackage{hyperref}
\usepackage{subfloat}
\usepackage{subfig}
\newlength{\wdth}

\newcommand \E {{\mathbb{E}}}

\newcommand\blfootnote[1]{%
	\begingroup
	\renewcommand\thefootnote{}\footnote{#1}%
	\addtocounter{footnote}{-1}%
	\endgroup
}

\begin{document}
	\title{Memorization in Deep Neural Networks: Does the Loss Function matter?}
	%
	%

	\author{Deep Patel \and P S Sastry}
	

	\authorrunning{Patel, D et al.}

	%
	
	\institute{Indian Institute of Science, Bangalore, India - 560012 \\
		\email{\{deeppatel,sastry\}@iisc.ac.in}}

	%
	\maketitle              
	\begin{abstract}
		Deep Neural Networks, often owing to the overparameterization, are shown to be capable of exactly memorizing even randomly labelled data. Empirical studies have also shown that none of the standard regularization techniques mitigate such overfitting. We investigate whether choice of loss function can affect this memorization. We empirically show, with benchmark data sets MNIST and CIFAR-10, that a symmetric loss function as opposed to either cross entropy or squared error loss results in significant improvement in the ability of the network to resist such overfitting. We then provide a formal definition for robustness to memorization and provide theoretical explanation as to why the symmetric losses provide this robustness. Our results clearly bring out the role loss functions alone can play in this phenomenon of memorization.
		
	\end{abstract}
	\section{Introduction}
	
	\blfootnote{Accepted at Pacific-Asia Conference on Knowledge Discovery and Data Mining (PAKDD-2021), New Delhi, India}Deep Neural Networks have been remarkably successful in a variety of classification problems involving image, text or speech data \cite{krizhevsky2017imagenet, yolo, elmo, wang2019structbert,raffel2020exploring}. 
	This is remarkable because these networks often have a large number of parameters and are trained on data sets that are not large enough for the sizes of these networks. This raises many questions about the (unreasonable) effectiveness of deep networks in  applications and whether they can go wrong on some kind of data sets. 
	
	In an interesting recent study, \cite{zhang} showed that standard deep network architectures are highly susceptible to extreme overfitting. They show that when one randomly alters class labels in the training data, these networks can learn the random labels almost exactly (with the gradient based learning algorithm driving the training error to near zero). It is seen that this memorization of the training examples cannot be mitigated through any of the standard regularization techniques such as weight decay or dropout. These results seem to imply that the usual complexity measures of statistical learning theory are inadequate to understand the learning dynamics of deep neural networks. In a further study, \cite{arpit} investigates this phenomenon more closely. While their study also confirms this memorization, they formulate some characterizations under which the learning dynamics of a network differ for the two cases of learning from real data and random data. Their study suggests that the data may be playing a vital role in resisting brute-force memorization by a network. While these studies experiment with many scenarios of regularization techniques and randomization of data, the role that the loss function itself can play in this has not been investigated. Motivated by this, here we present some experiments to show that a loss function can also play a significant role in preventing a network from memorizing data. 
	
	Neural networks are universal approximators \cite{nn-approx-3} and  networks with sufficient parameters have the capacity to exactly represent any finite amount of data \cite{zhang}. Such results show that there exist parameter values that can represent any arbitrary function. However, as discussed in \cite{arpit}, what a network learns depends on the parameter values that a gradient-based learning algorithm can reach starting from some random initial parameter values. This learning dynamics is certainly affected, among other factors, by the loss function because the loss function determines the topography of the empirical risk, which is minimized by the learning algorithm. Hence, it would be interesting to investigate whether it is possible to have loss functions that can inherently resist (to some degree) the memorization of data by a network. 
	
	Here we present some experimental results for benchmark datasets, MNIST \cite{mnist} and CIFAR-10 \cite{cifar10}, with labels randomly changed with different probabilities. We see that for varying probabilities of random labelling, networks trained with standard loss functions such as categorical cross entropy (CCE) or mean square error (MSE) exhibit memorization by reaching close to zero training error. Then, we investigate learning these same networks using a special class of loss functions -- \textit{symmetric loss functions}. We specifically use the so-called \textit{robust log loss} (RLL) \cite{rll} -- which is obtained by modifying CCE -- though we also comment on other similar loss functions. We show that keeping everything else in the training algorithm same but changing the loss function alone results in the network significantly resisting overfitting. With these loss functions the training error saturates at a level much above zero (depending on the amount of random label flipping). We also see that the learning dynamics with these symmetric loss functions resembles more of what one expects with the clean, real data. 
	It was suggested in \cite{arpit} that, with real data, the networks try to fit the patterns in the data rather than memorizing the data, while with randomly flipped labels the networks seem to be using brute-force memorization. We show that with fairly high (though less than 100\%) randomization of labels in training data, networks trained with CCE or MSE loss seem to be using brute-force memorization while the same networks trained with the symmetric loss, RLL, seem to be resisting such memorization by trying to fit mainly the clean part of the data. This adequately demonstrates that the loss function also has an important role to play in resisting this type of memorization of data. 
	
	We also present some theoretical justification (using the known properties of these symmetric losses) for the ability of these loss functions to resist overfitting. For the case of random label flipping, we formally define what can be called \textit{resisting of overfitting} or \textit{memorization}. Using this, we explain why symmetric loss functions can resist brute-force memorization in these scenarios. The analysis we present provides some theoretical justification for the empirically observed performance with RLL. We discuss the implications of this and speculate on how loss functions may be crucial in realizing better learning dynamics.
	
	\subsection{Related Work} 
	Memorization in deep networks got a lot of attention recently due to  \cite{zhang} which showed that SGD-based training of neural networks drives the training set accuracy to 100\% even in case of randomly labelled data with none of the standard regularization methods being helpful for avoiding this memorization. They speculate on the implications of this for characterizing the generalization abilities of networks. 
	In further studies, \cite{arpit,memorization-3} characterize the behaviour of neural networks on real and randomly-labelled data experimentally and find that deep networks learn simpler patterns first before starting to memorize the data. They also claim that explicit regularization such as dropout can actually help resist memorization to some extent.\cite{memorization-4} shows that memorization is necessary for generalization for some types of distributions which has been tested empirically by \cite{memorization-5}. However, none of these studies investigate whether the loss function has a role in memorization and that is what is explored in this paper
	
	There are many works that attempt comparative study of loss functions for classification tasks. \cite{cce-vs-mse} shows, with extensive empirical experiments on a variety of data sets, that MSE performs better than CCE thus challenging the conventional wisdom of the superiority of CCE loss for classification tasks. \cite{cce-vs-mse-2} argue that CCE is favourable (compared to MSE) for multi-class settings but propose a technique that makes performance of MSE comparable to that of CCE. \cite{mse-vs-hinge, mse-vs-hinge-2} find MSE has comparable or better performance than hinge loss for several tasks. \cite{loss-fn-matter} show that minimizers of risk obtained in case of MSE and hinge loss are the same for overparameterized linear models under certain conditions. These and other similar works compare different loss functions for classification and regression tasks from the point of view of generalization whereas our work looks at the role loss functions can play in affecting the degree of memorization in overparameterized networks.
	
	The problem of learning under label noise, that is, learning when training data has random labeling errors, has also been extensively studied in recent years. (See, e.g., \cite{naresh2013,frenay-survey,ghosh2017aaai,meta-net,loss-correction}).
	In tackling label noise the focus is mostly on algorithms that deliver good performance by, e.g., sample reweighting, label cleaning, loss correction, etc. In this work our focus is on the inherent robustness of a loss function and not on any algorithmic modifications to take care of label noise. 

	The main contributions of the paper are as follows: We consider some scenarios of network architectures and randomization of training labels under which deep networks are earlier demonstrated to be susceptible to memorization. We show through empirical studies that training the same network with a different loss function, namely, RLL, can significantly resist this memorization as compared to training with standard CCE or MSE. Our experiments adequately demonstrate that the loss function has a crucial role and supports our viewpoint that it is important to study such properties of loss functions. We propose a formal definition for the ability of a network to resist overfitting of the kind studied \cite{zhang,arpit}. Using this definition, for these scenarios of random label flipping on training data, we provide theoretical justification for the observed performance with the symmetric loss functions.
	
	The rest of the paper is organized as follows: In Section \ref{section:2}, we present our empirical studies with the CCE, MSE, \& RLL loss functions. Section \ref{section:3} presents our theoretical analysis. Conclusions are presented in Section \ref{section:4}.
	
	
	\section{Role of loss function in resisting memorization}
	\label{section:2}
	
	We experiment with two network architectures. One is an Inception-like network architecture (referred to as Inception-Lite in this paper) which is same as that used in \cite{zhang} for demonstrating memorization in deep networks. The second is ResNet-32 (and ResNet-18 for MNIST) architecture as used in \cite{meta-net}. 
	
	In this section we present results with three loss functions. Two are the standard loss functions used with neural networks, namely, CCE and MSE, and the third is a symmetric loss, viz. RLL. Since we are considering classification problems, for all the networks we assume a softmax output layer. For an input, $\mathbf{x}$, let $\mathbf{g}(\mathbf{x})$ denote the vector output of the network with components $g_i(\mathbf{x})$.   When $\mathbf{x}$ belongs to class $k$, the label would be the one-hot vector $\mathbf{e}^k$ where $e^k_k=1$ and $e^k_j = 0, \; \forall j \neq k$. Let $K$ denote the number of classes. With this notation, the three loss functions can be defined as follows:
	
	\noindent
	\begin{eqnarray*}
		\mathcal{L}_{CCE}(\mathbf{g}(\mathbf{x}), \mathbf{e}^k) & = & -\sum_i e^k_i \; \log\left(g_i(\mathbf{x})\right) \;= -\log(g_k(\mathbf{x}))  \\
		\mathcal{L}_{MSE}(\mathbf{g}(\mathbf{x}), \mathbf{e}^k) & = & \sum_i \left(g_i(\mathbf{x}) - e^k_i\right)^2\\
		\mathcal{L}_{RLL}(\mathbf{g}(\mathbf{x}), \mathbf{e}^k) & = & \log\left(\frac{\alpha+1}{\alpha}\right) - \log(\alpha+g_k(\mathbf{x})) + \sum_{j\neq k} \frac{1}{K-1} \log(\alpha + g_j(\mathbf{x}))
	\end{eqnarray*}
	where $\alpha > 0$ is a parameter of the RLL. 
	
	We can get some insights on behaviour of RLL versus CCE as follows: When $\mathbf{x}$ is in class-$k$, $g_k(\mathbf{x})$ is the posterior probability assigned to class-$k$ by the network. If this is high, then the CCE loss, which is $-\log(g_k(\mathbf{x})$, is low. However, the CCE loss is unbounded because, in principle, $g_k(x)$ can be arbitrarily small. Disregarding the constant term, the RLL takes $-\log(\alpha+g_k(\mathbf{x})) + \sum_{j\neq k} \frac{1}{K-1}\log(\alpha+g_j(\mathbf{x}))$ as its value. Since we are using $\log(\alpha+g_j(\mathbf{x}))$ rather than $\log(g_j(\mathbf{x}))$, the loss is now bounded. More importantly, the loss is essentially determined through a kind of comparison of the posterior probability assigned to class-$k$ by the network against the average probability assigned to all other classes. (The constant term in RLL is there only to ensure that the loss is non-negative). As we shall see, this gives some amount of robustness in the risk minimization resulting in RLL exhibiting good resistance to memorization. 
	
	We train all the networks to minimize empirical risk (with each of the loss functions). We employ mini-batch based stochastic gradient descent (SGD) for Inception-Lite \& ResNet-32 (for CIFAR-10) and Adam \cite{adam} for ResNet-18 (for MNIST). For Inception-Lite, we use a constant step-size of 0.01 in each epoch which is reduced by a factor of $0.95$ after each epoch for 100 epochs whereas a constant step-size of 0.1 is used for ResNet-32 which is reduced by a factor of 0.1 after 100 and 150 epochs. ResNet-32 \& ResNet-18 are trained for 200 epochs. The ResNet-18 is trained with a step-size of 0.001. Inception-Lite is trained for 100 epochs because the training accuracies saturate by then. Inception-Lite and ResNet-18 are trained without weight decay whereas ResNet-32 is trained with a weight decay of 0.0001. 
	
	CIFAR-10 and MNIST benchmark datasets are used for the experiments. As explained earlier, we study the memorization by the networks through randomly altering the class labels in the training set. For this, independently for each example, we retain the original label with probability $(1-\eta)$ and change it with probability $\eta$. When the label is changed, it is changed to one of the other classes with equal probability. We experiment with $\eta = 0, 0.2, 0.4,$ and $0.6$. (Note that $\eta=0$ corresponds to the clean or original training data). By varying $\eta$ we can change the amount of pattern information present in training data and hence can study whether a loss function can result in learning this information. Here, we are considering 10-class classification task. Our randomization of labels is such that up to $\eta < 0.9$, in an expectation sense, for any class-$j$, the number of data points in the training set that are correctly labelled as class-$j$ would be more than the number of data points of a class-$i$, $i \neq j$, incorrectly labelled as class-$j$. Hence, at $\eta$ well below $0.9$ there should be scope for learning the underlying patterns and not overfitting the randomized training data.
	
	\begin{figure}[ht]
		\centering
		\subfloat[$CCE$]{\label{fig:cifar-resnet-cce-train-acc}\includegraphics[trim= 10mm 0mm 22mm 0mm, width=30mm, height=30mm]{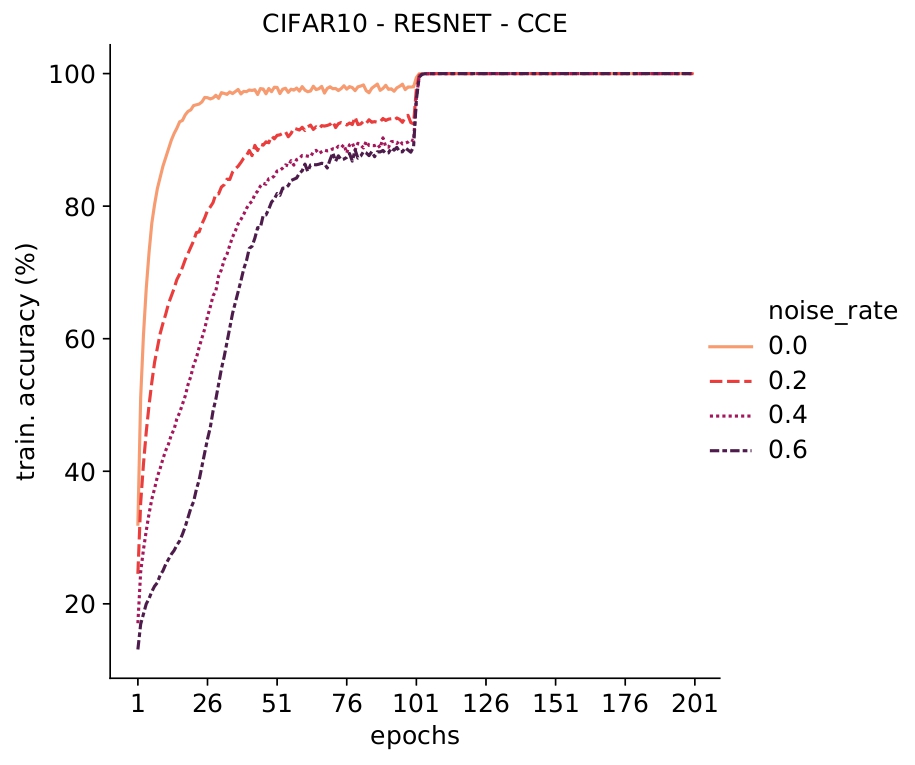}}
		\subfloat[$MSE$]{\label{fig:cifar-resnet-mse-train-acc}\includegraphics[trim= 10mm 0mm 22mm 0mm, width=30mm, height=30mm]{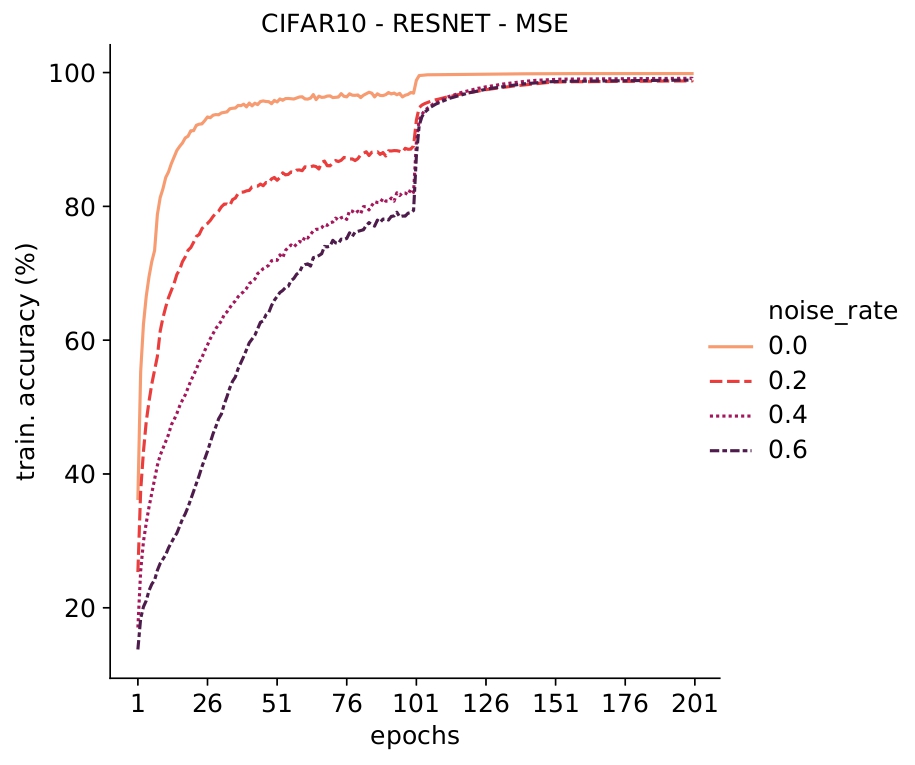}}
		\subfloat[$CCE$]{\label{fig:cifar-inception-cce-train-acc}\includegraphics[trim= 10mm 0mm 22mm 0mm, width=30mm, height=30mm]{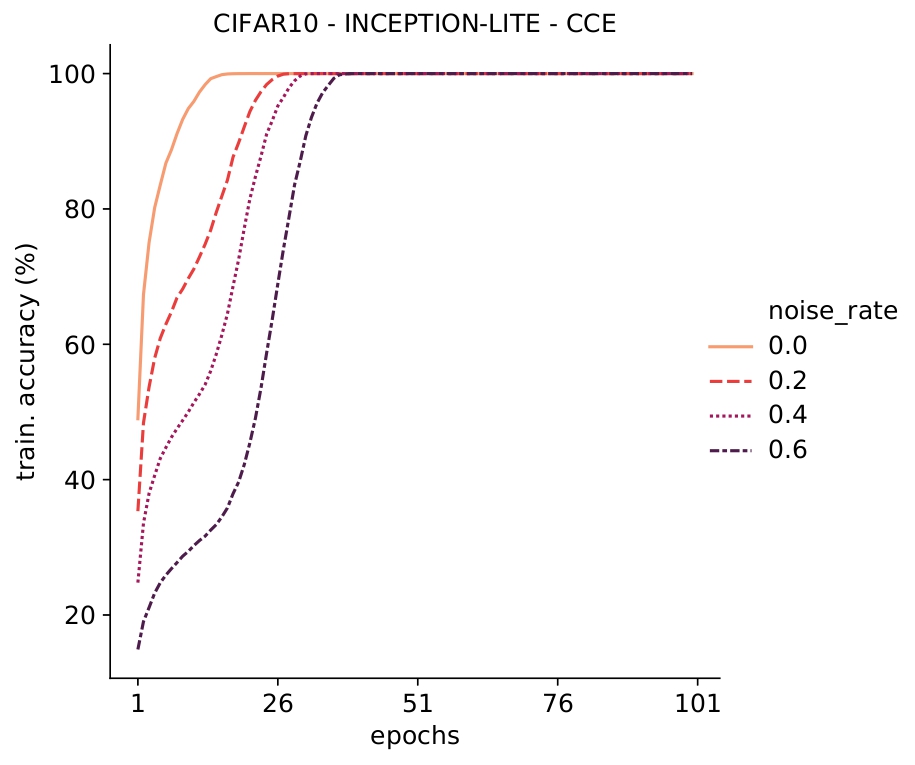}}
		\subfloat[$MSE$]{\label{fig:cifar-inception-mse-train-acc}\includegraphics[trim= 10mm 0mm 25mm 0mm, width=30mm, height=30mm]{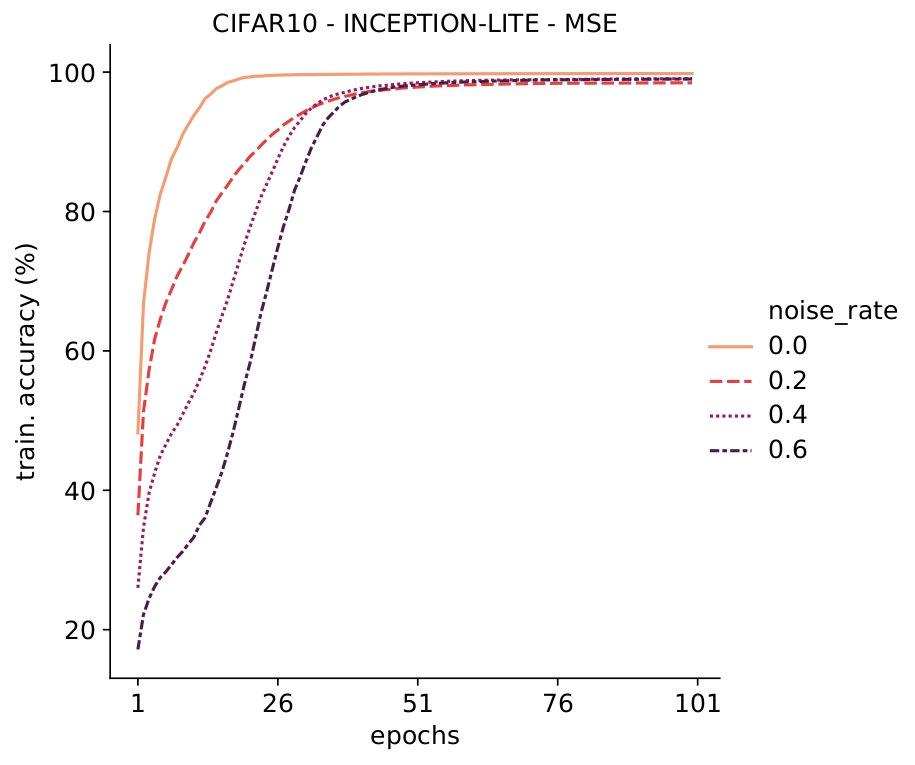}}

		
		\caption{Training set accuracies for ResNet-32 (\protect\subref{fig:cifar-resnet-cce-train-acc} \& \protect\subref{fig:cifar-resnet-mse-train-acc}) \& Inception-Lite (\protect\subref{fig:cifar-inception-cce-train-acc} \& \protect\subref{fig:cifar-inception-mse-train-acc}) trained on CIFAR-10 with CCE and MSE losses for for $ \eta \in \{0., 0.2, 0.4, 0.6\}$}
		\label{fig:cifar-resnet-inception-train-acc}
		
	\end{figure}
	
	\begin{figure}[ht]
		\centering
		
		\subfloat[$CCE$]{\label{fig:mnist-resnet-cce-train-acc}\includegraphics[trim= 10mm 0mm 22.8mm 0mm, width=40mm, height=30mm]{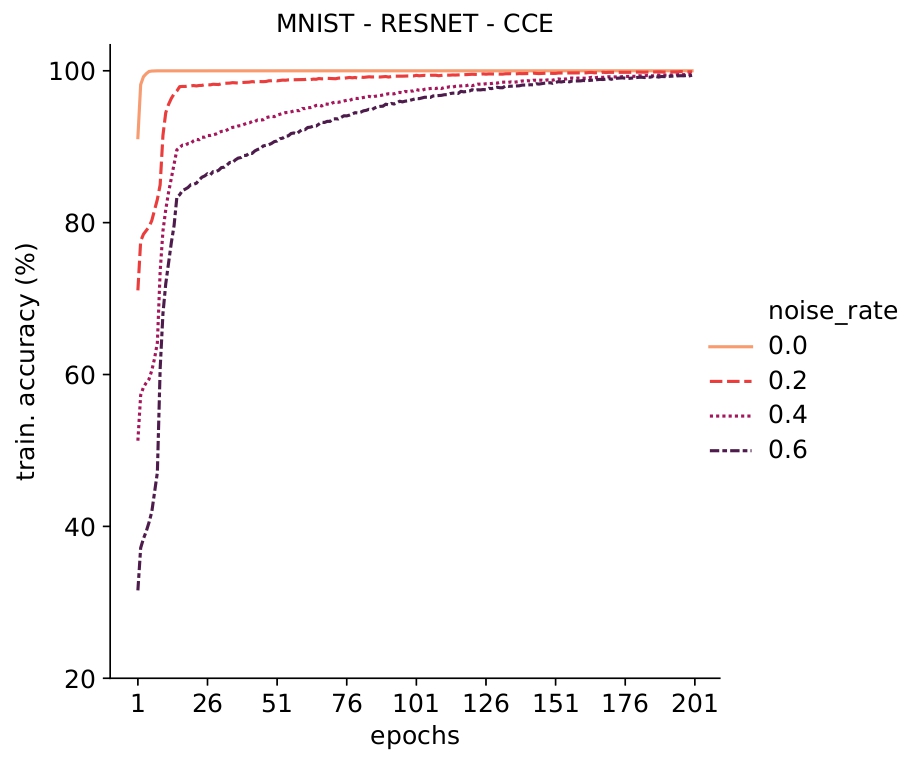}}
		\subfloat[$MSE$]{\label{fig:mnist-resnet-mse-train-acc}\includegraphics[trim= 10mm 0mm 25mm 0mm, width=40mm, height=30mm]{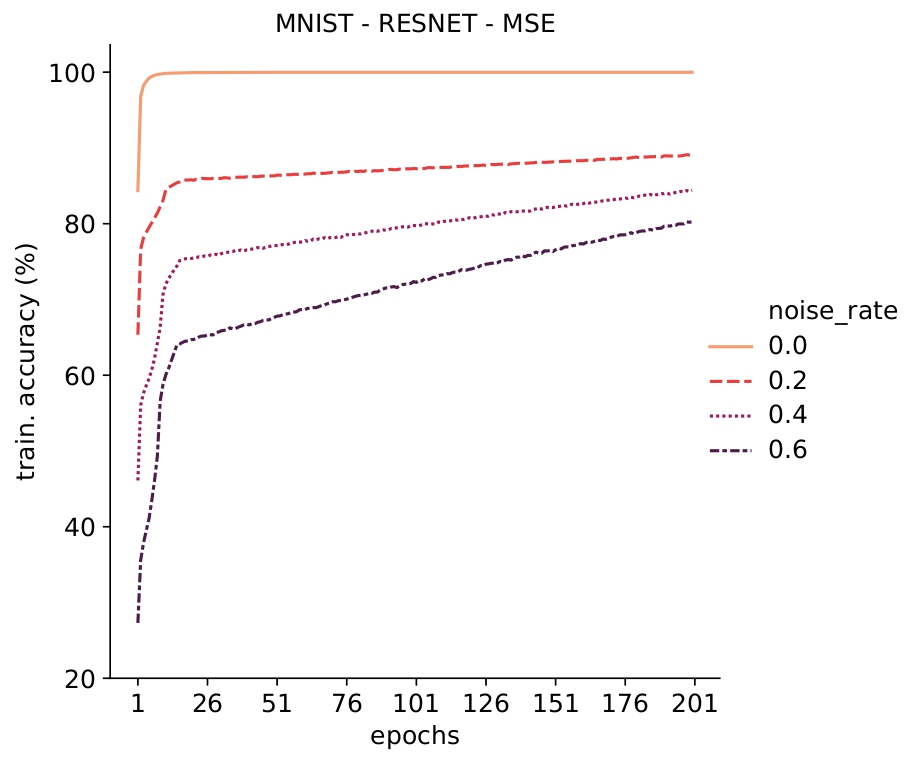}}
		
		\caption{Training set accuracies for ResNet-18 trained on MNIST with CCE and MSE losses for different levels of label noise}
		\label{fig:mnist-resnet-train-acc}

	\end{figure}
	
	\begin{figure}[ht]
		\centering
		\subfloat[$ResNet-32$]{\label{fig:cifar-resnet-rll-train-acc}\includegraphics[trim= 10mm 0mm 34mm 0mm, width=35mm, height=30mm]{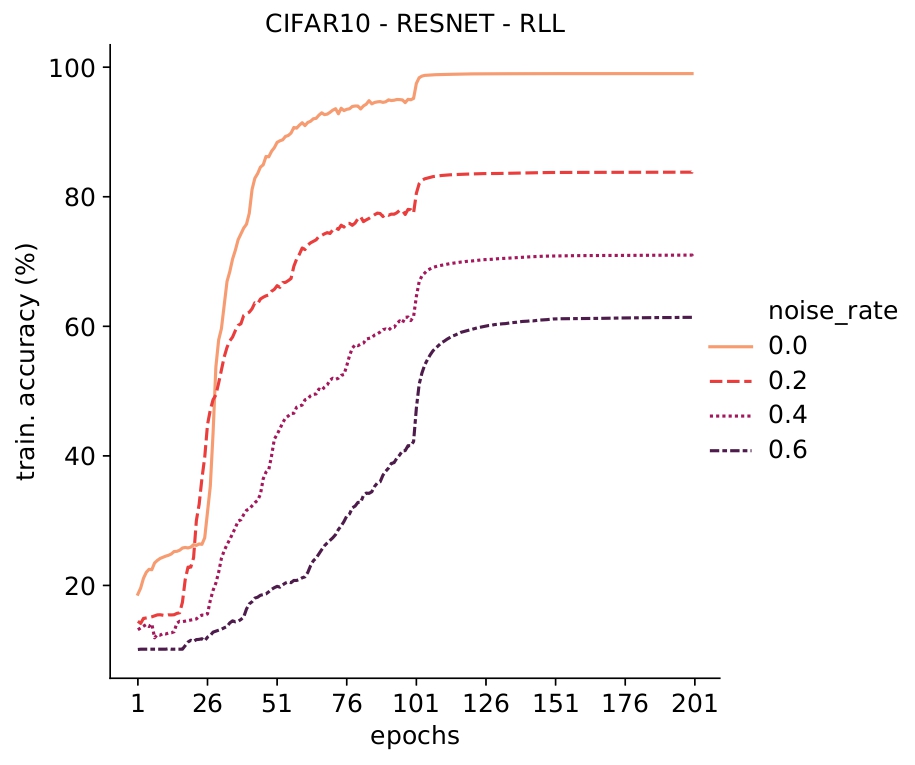}}
		\subfloat[$ResNet-18$]{\label{fig:mnist-resnet-rll-train-acc}\includegraphics[trim= 0mm 0mm 34mm 0mm, width=35mm, height=30mm]{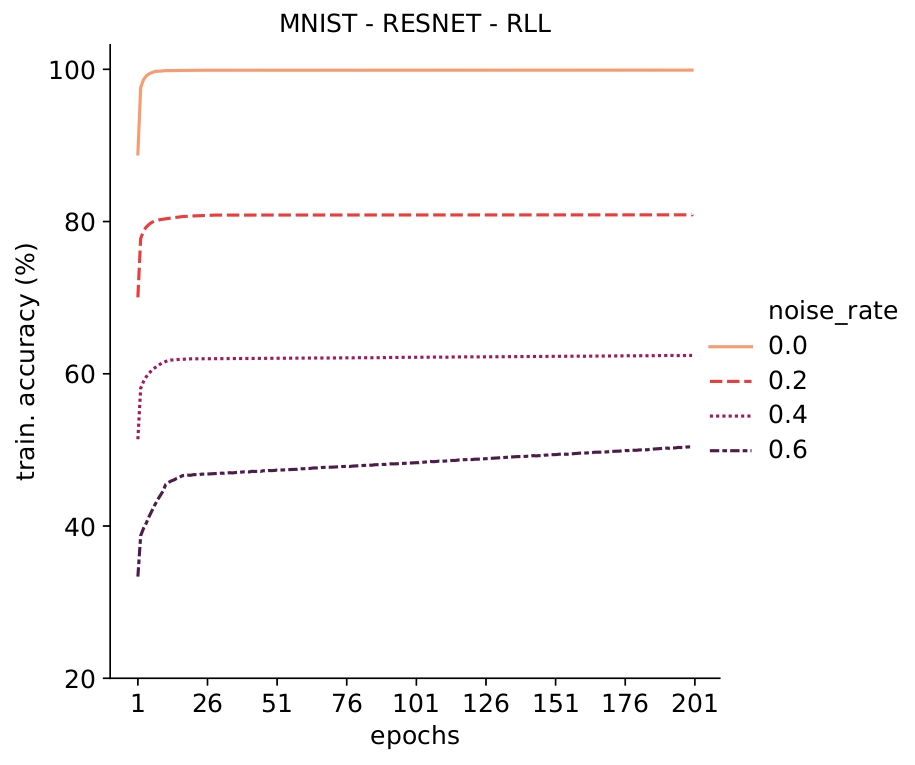}}
		\subfloat[$Inception-Lite$]{\label{fig:cifar-inception-rll-train-acc}\includegraphics[trim= 0mm 0mm 34mm 0mm, width=35mm, height=30mm]{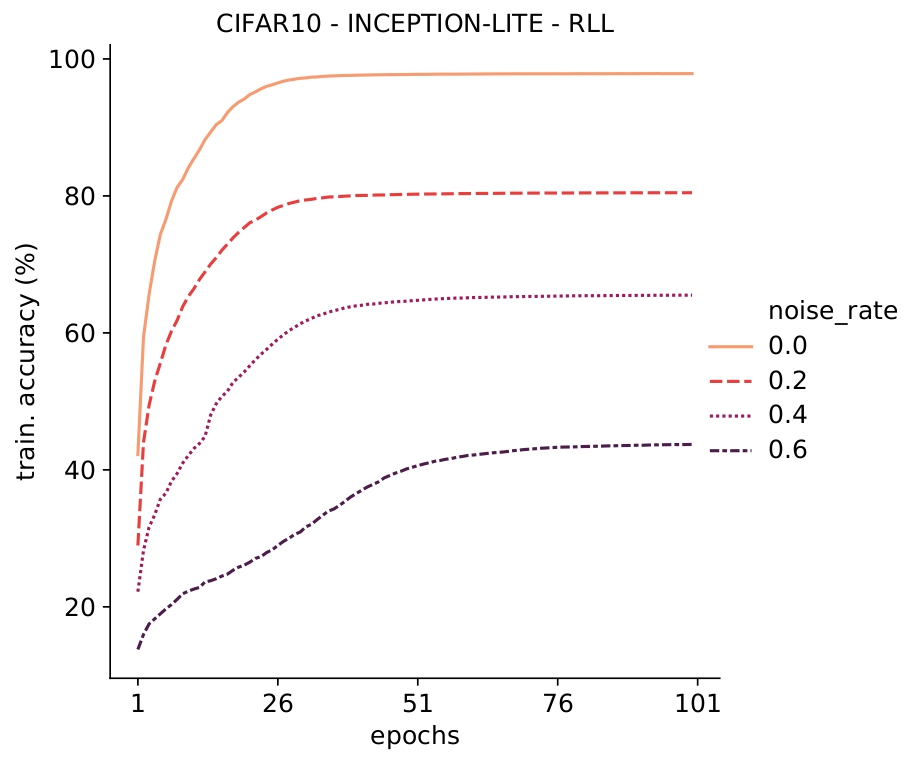}}
		
		\caption{Training set accuracies for networks trained on CIFAR-10 (\protect\subref{fig:cifar-resnet-rll-train-acc} \& \protect\subref{fig:cifar-inception-rll-train-acc}) \& MNIST (\protect\subref{fig:mnist-resnet-rll-train-acc}) with RLL for $ \eta \in \{0., 0.2, 0.4, 0.6\}$}
		\label{fig:cifar-mnist-train-acc-rll}
		
	\end{figure}
	
	
	Figure \ref{fig:cifar-resnet-inception-train-acc} shows the training accuracies achieved with ResNet-32 and InceptionLite when we train the network with CCE \& MSE for various values of $\eta$ on CIFAR-10 while Figure \ref{fig:mnist-resnet-train-acc} shows training accuracies of ResNet-18 with CCE and MSE for MNIST. As can be seen from the figures, for all values of $\eta$ the training error goes down close to zero though it takes a few epochs more with higher values of $\eta$. The only exception is when ResNet-18 is trained on MNIST with MSE; but even here the training accuracy reaches a high value.  This is consistent with the results reported in \cite{zhang,arpit}. (Note that \cite{zhang} show training set performances only for $\eta=1$ and do not experiment with varying levels of noise as was done here.) Note that at $\eta=0.2$, 80\% of training samples of a class are correctly labelled and hence would contain the patterns that the network would have learnt when trained with clean data. However, the network ends up learning a function that can exactly reproduce the training set labels. This seems to indicate that with these loss functions the topography of the empirical risk function is such that the learning dynamics takes the network to a point that fits the random labels exactly. The brute-force memorization manifests itself in these networks trained with CCE  even at moderate levels of label randomization.
	
	These results may be contrasted with those presented in Figure \ref{fig:cifar-mnist-train-acc-rll} which are obtained when the same networks are trained with RLL for different values of $\eta$ on MNIST \& CIFAR-10. As can be seen from the figures, the training set accuracy achieved by RLL for non-zero values of $\eta$ is always well below that achieved on clean data. This shows that the network does not blindly learn to reproduce the training set labels. This is significant because this shows that when we keep everything else same and change only the loss function, the learning dynamics now seem to be able to resist brute-force memorization.  Also, for $\eta=0.2$ and $\eta=0.4$ the difference in training-accuracy on clean and noisy data is almost equal to the noise-rate thus suggesting that this loss function seems to be able to disregard data that are wrongly labelled. 
	
	We now take a closer look to understand the kind of classifier learnt by RLL under noisy data. Let $\{X_i, y_i\}_{i=1}^{\ell}$ denote the original training data and let $\{X_i, \tilde{y_i}\}_{i=1}^{\ell}$ denote the noisy or randomly-labelled data given to the learning algorithm. Let $h(X)$ denote the actual class label predicted by the network for $X$ (which is determined by $\max(g_i(X))$ where $g(X)$ is the output of softmax layer).  Then the training accuracy, say $J_1$, is defined by
	\[J_1 = \frac{1}{\ell}\sum_{i=1}^\ell I_{[h(X_i) = \tilde{y}_i)]} \]
	where $I_A$ is indicator of $A$. This is the accuracy defined with respect to the labels as given in the training set. We define another accuracy, $J_2$, by
	\[J_2 = \frac{1}{\ell}\sum_{i=1}^\ell I_{[h(X_i) = y_i)]} \]
	$J_2$ is the accuracy with respect to original, uncorrupted training set. This accuracy indicates how well the network, learned with randomly-altered labels, would be able to reproduce the original clean labels of the training data. 
	
	\begin{figure}[ht]
		\centering
		\subfloat[$CCE$]{\label{fig:cifar-inception-cce-j1}\includegraphics[trim= 10mm 0mm 34mm 0mm, width=30mm, height=30mm]{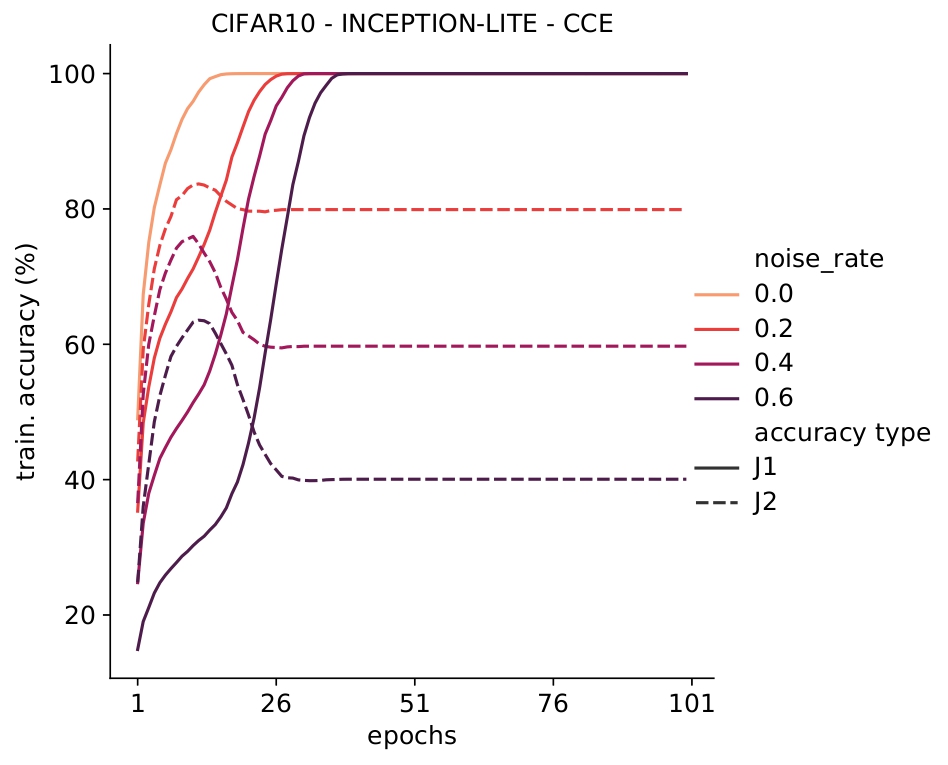}}
		\subfloat[$RLL$]{\label{fig:cifar-inception-rll-j1}\includegraphics[trim= 6.5mm 0mm 34mm 0mm, width=30mm, height=30mm]{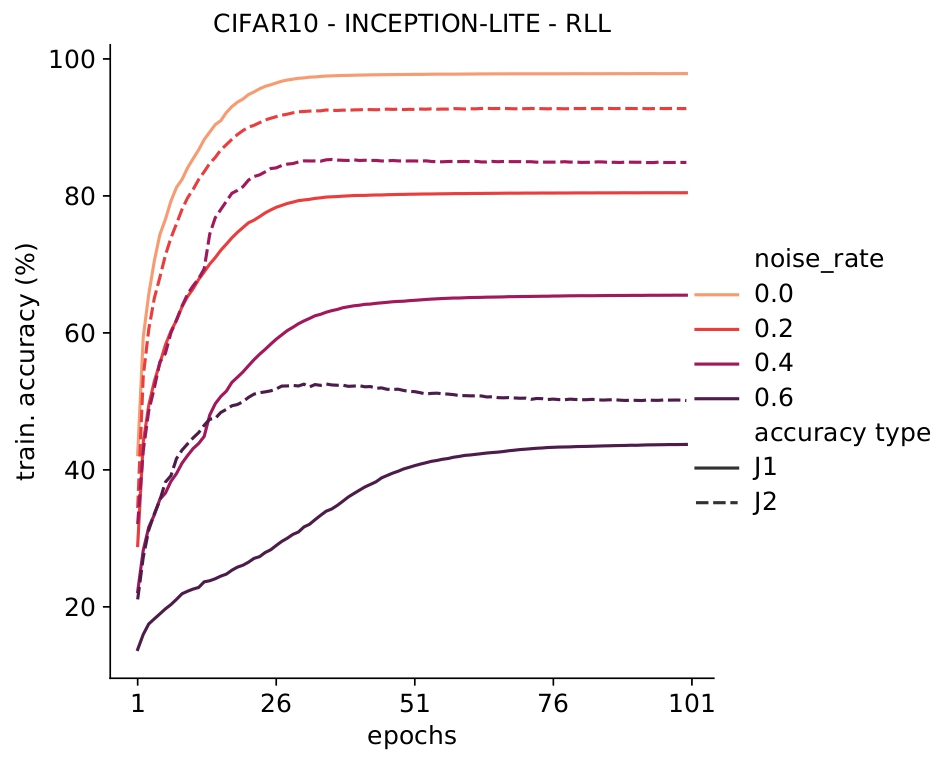}}
		\subfloat[$MSE$]{\label{fig:mnist-resnet-mse-j1}\includegraphics[trim= 6.5mm 0mm 34mm 0mm, width=30mm, height=30mm]{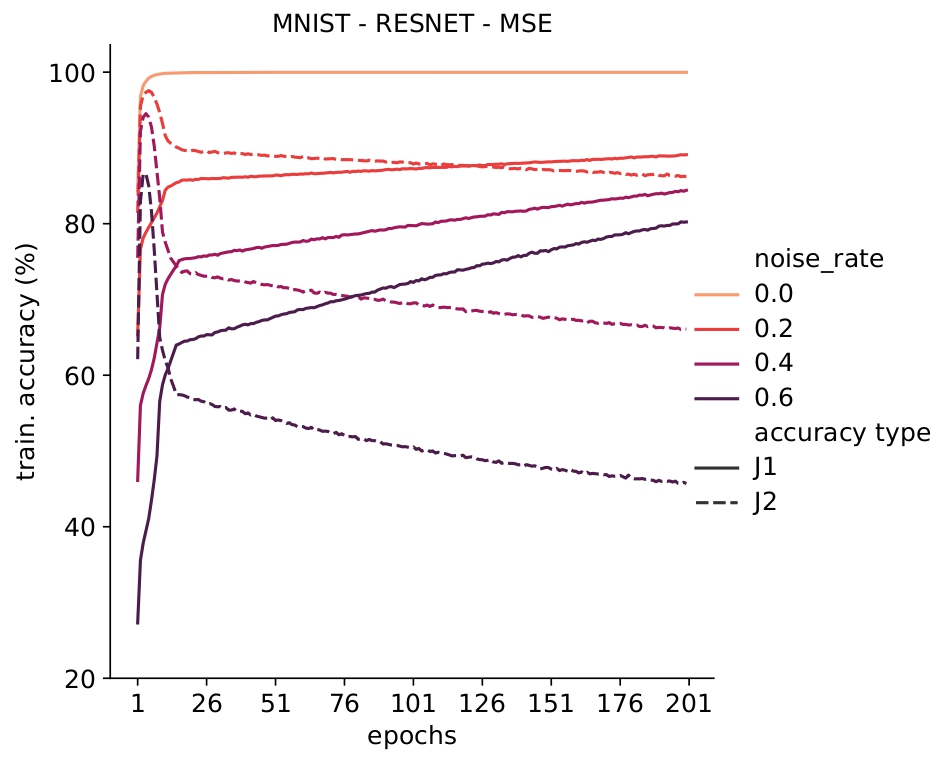}}
		\subfloat[$RLL$]{\label{fig:mnist-resnet-rll-j1}\includegraphics[trim= 6.5mm 0mm 25mm 0mm, width=30mm, height=30mm]{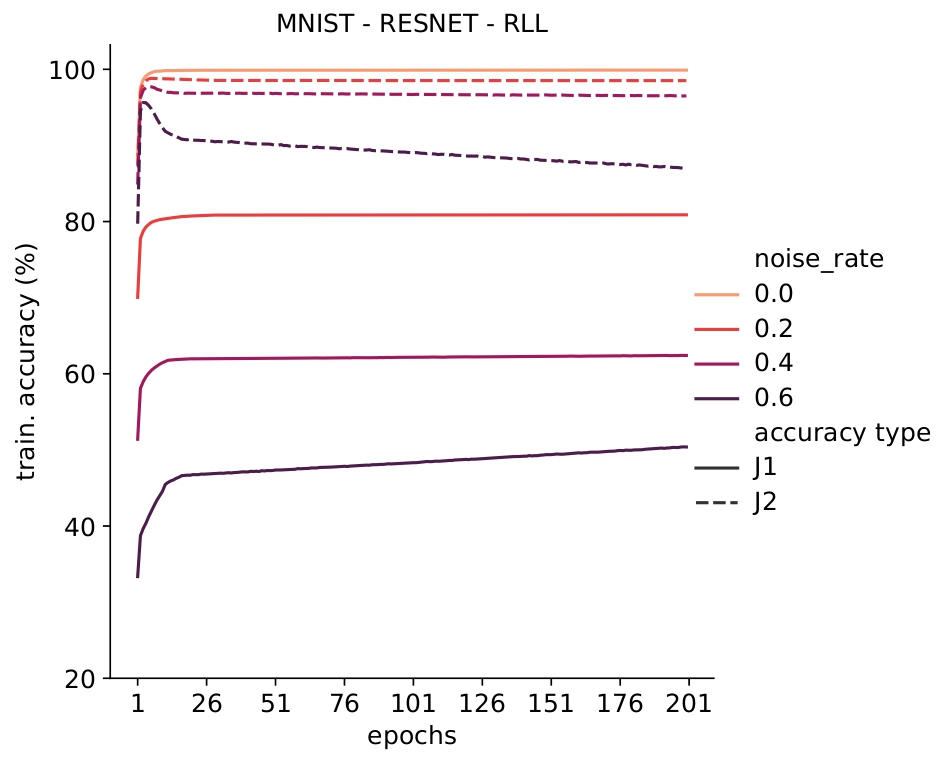}}
		
		\caption{$J_1$ and $J_2$ accuracies for Inception-Lite (\protect\subref{fig:cifar-inception-cce-j1} \& \protect\subref{fig:cifar-inception-rll-j1}) \& ResNet-18 (\protect\subref{fig:mnist-resnet-mse-j1} \& \protect\subref{fig:mnist-resnet-rll-j1}) trained on CIFAR-10 and MNIST resp. for $ \eta \in \{0., 0.2, 0.4, 0.6\}$ (Solid lines show $J_1$ accuracy; dashed lines show $J_2$ accuracy)}
		\label{fig:cifar-mnist-resnet-j1}
		
	\end{figure} 
	
	\begin{figure}[ht]
		\centering
		\subfloat[$Norm. MSE$]{\label{fig:cifar-inception-norm-mse-train}\includegraphics[trim= 10mm 0mm 25mm 0mm, width=30mm, height=30mm]{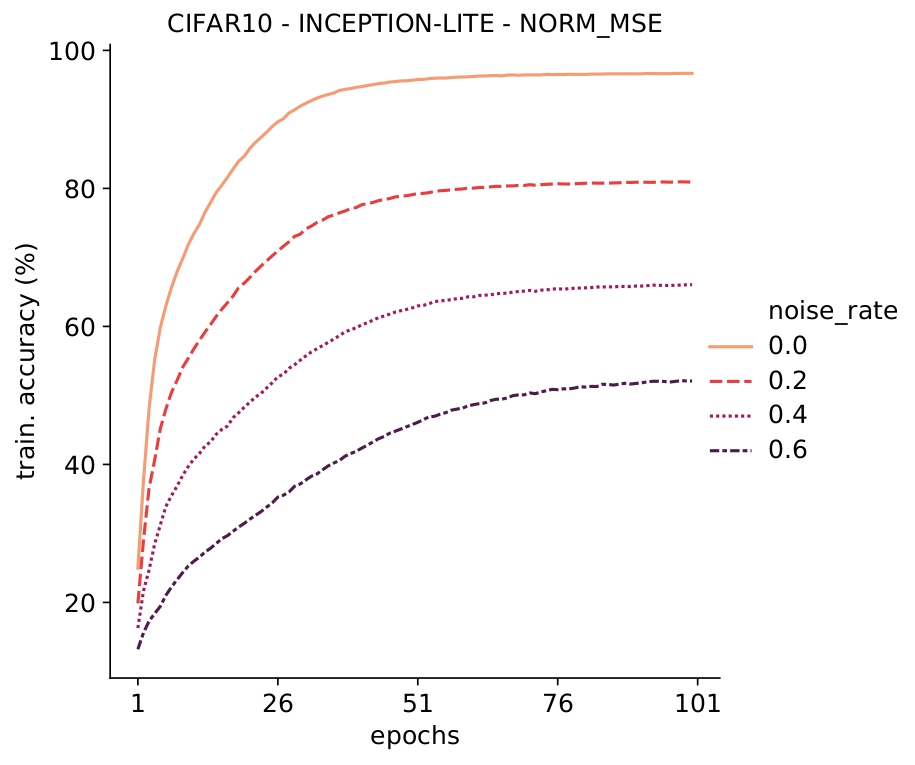}}
		\subfloat[$Norm. MSE$]{\label{fig:mnist-resnet-norm-mse-train}\includegraphics[trim= 10mm 0mm 5mm 0mm, width=30mm, height=30mm]{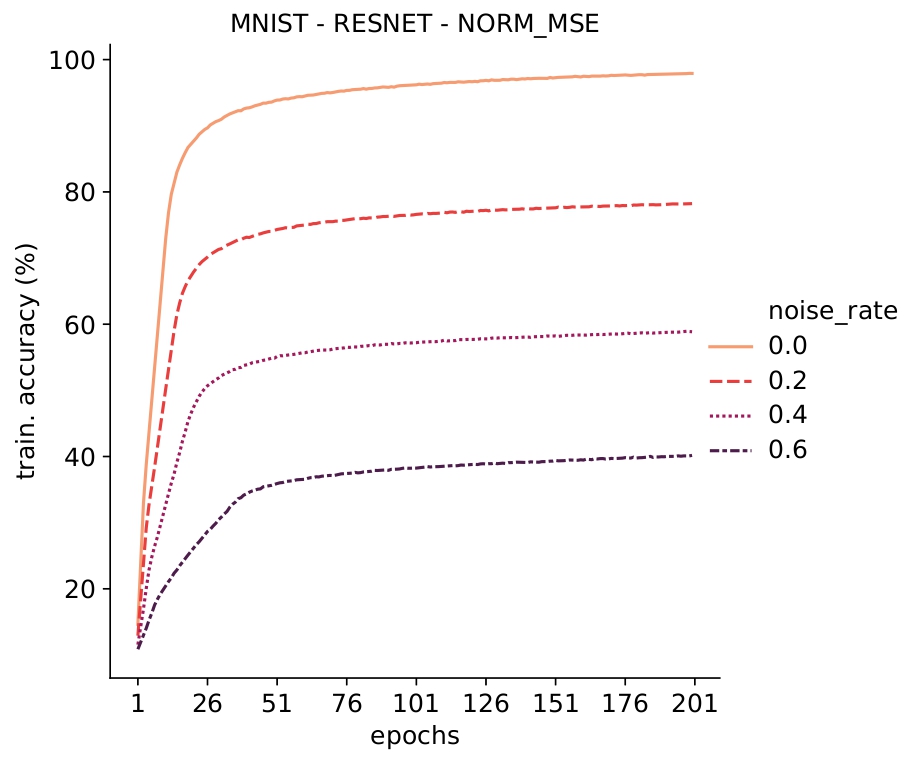}}
		\subfloat[$Norm. MSE$]{\label{fig:cifar-inception-norm-mse-j1}\includegraphics[trim= 5mm 0mm 30.5mm 0mm, width=30mm, height=30mm]{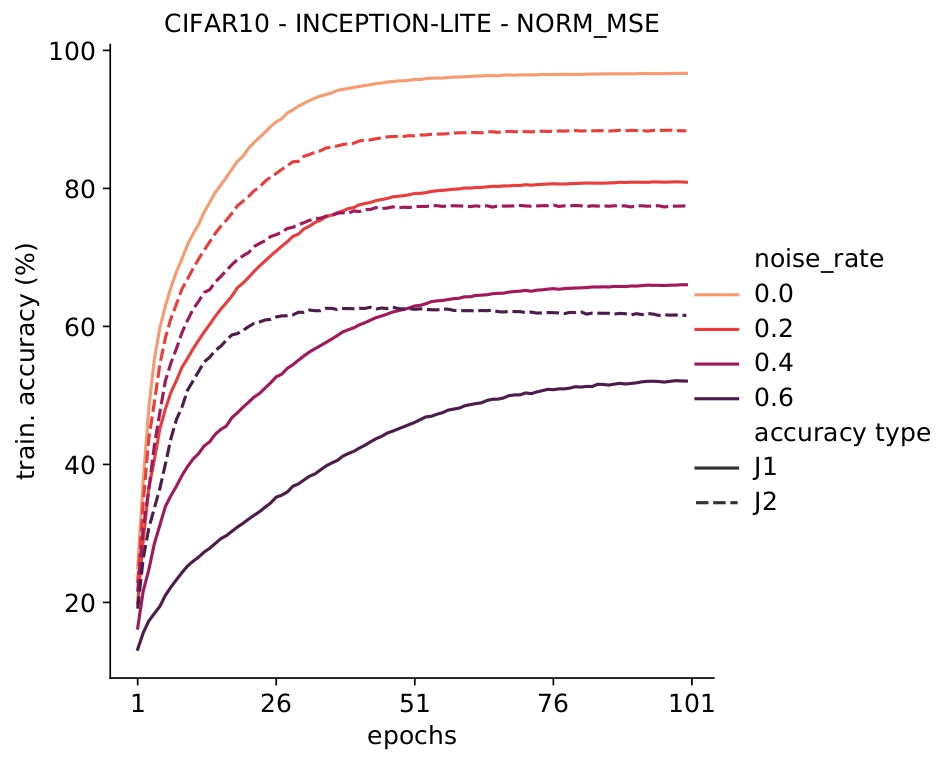}}
		\subfloat[$Norm. MSE$]{\label{fig:mnist-resnet-norm-mse-j1}\includegraphics[trim= 10mm 0mm 25mm 0mm, width=30mm, height=30mm]{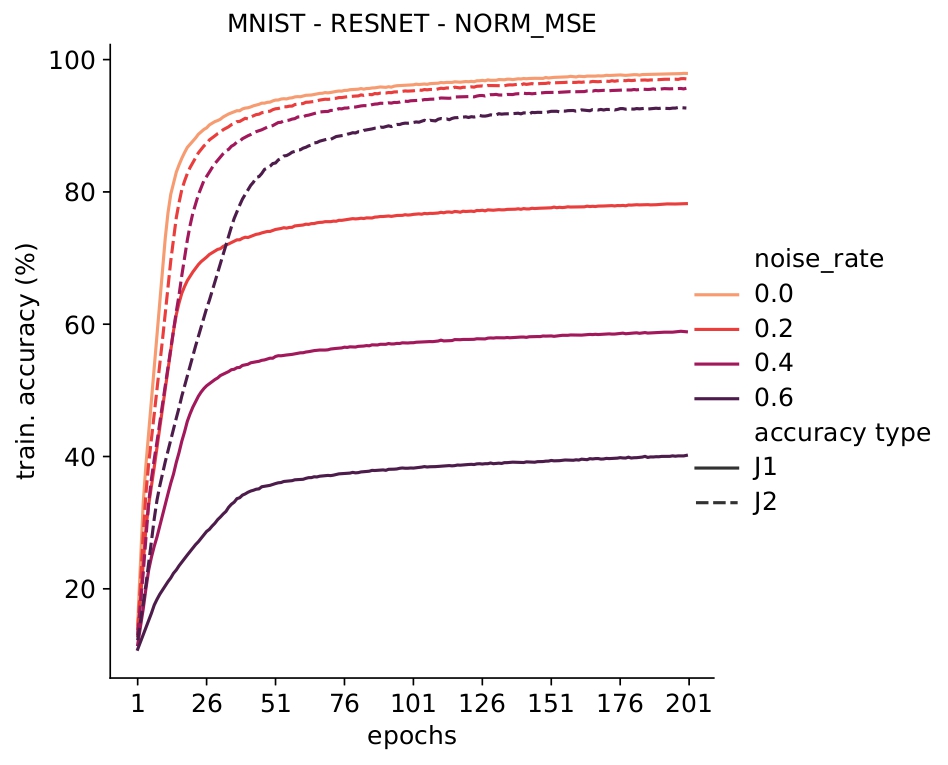}}
		
		\caption{Train. accuracy and $J_1$ \& $J_2$ accuracies for Inception-Lite (\protect\subref{fig:cifar-inception-norm-mse-train} \& \protect\subref{fig:cifar-inception-norm-mse-j1}) \& ResNet-18 (\protect\subref{fig:mnist-resnet-norm-mse-train} \& \protect\subref{fig:mnist-resnet-norm-mse-j1}) trained on CIFAR-10 and MNIST resp. for $ \eta \in \{0., 0.2, 0.4, 0.6\}$  (Solid  lines show $J_1$ accuracy; dashed lines show $J_2$ accuracy)}
		\label{fig:cifar-mnist-norm-mse}
		
	\end{figure} 
	
	We show in Figures \ref{fig:cifar-mnist-resnet-j1} the accuracies $J_1$ and $J_2$ for networks learned with the different loss functions for different values of $\eta$. As can be seen from Figures \ref{fig:cifar-inception-cce-j1} \& \ref{fig:mnist-resnet-mse-j1} for networks trained with CCE and MSE losses, the $J_2$ accuracy (dashed line) is always well below the $J_1$ accuracy (solid line). This is as expected because, as seen earlier, the training accuracy, which is equal to $J_1$, is close to 100\%. However, for networks learned with RLL (Figure \ref{fig:cifar-inception-rll-j1} \& \ref{fig:mnist-resnet-rll-j1}), it is the $J_2$ (dashed line) accuracy that is always higher than the $J_1$ accuracy (solid line). As a matter of fact, for $\eta=0.2, 0.4$, the $J_2$ accuracy of the networks learned using RLL is close to the training accuracy achieved with clean data. This suggests that this loss function is able to disregard the randomly altered labels and help the network learn a classifier that it would have learned with clean data.
	

	There is another interesting point about this figure. The figure shows how the $J_1$ and $J_2$ accuracies evolve with epochs. As can be seen from the figure, the networks learned using CCE with noisy data seemed to have initially tried to learn the patterns and thus the $J_2$ accuracy is higher in the early epochs. But eventually the network `flips' and overfits to the random labels in training data. However, this `flip' never happens for networks trained using RLL; through all the epochs, the $J_2$ accuracy stays higher.
	
	All the empirical results presented in this section amply demonstrate that a loss function can play a significant role in mitigating the memorization effect observed with deep neural networks. In the next section, we present some theoretical analysis that explains, to some extent, the results presented in this section. 
	
	\section{Robustness of Symmetric Loss Functions}
	\label{section:3}
	
	In~\cite{zhang}, for networks learned using training data with random labels, the accuracy obtained on part of the original data is taken as test error for the purpose of discussing the generalization abilities. However, this may be somewhat of an inaccurate nomenclature. Normally the test error is error on new data but drawn from the same distribution as that from which training data is drawn. 
	
	We will now present another way of formalizing this. For this section we assume class labels, $y_i$, take values in $\mathcal{Y} =\{1, \cdots, K\}$ rather than being one-hot vectors. Let $S=\{X_i, y_i\}_{i=1}^{\ell}$ be the original training data and we assume it is drawn {\em iid} according to a distribution $\mathcal{D}$. The training data with randomly altered labels is denoted by $S_{\eta}=\{X_i, \tilde{y_i}\}_{i=1}^{\ell}$, where, for each $i$, 
	\[\tilde{y}_i = \left\{ \begin{array}{ll}
	y_i & \mbox{~with probability~~} 1 -\eta \\
	j \in \mathcal{Y} - \{y_i\} & \mbox{~with probability~~} \frac{\eta}{K-1} 
	\end{array} \right. \]
	That is, $\tilde{y}_i$ is same as $y_i$ with probability $(1-\eta)$ and takes each of the other possible labels with equal probability. We denote the distribution from which $S_{\eta}$ is drawn as $\mathcal{D}_{\eta}$ and it is related to $\mathcal{D}$ as given above. 
	
	When one is empirically investigating memorization of random labels, one is using training data drawn according to distribution $\mathcal{D}_{\eta}$ but is interested in test error according to distribution $\mathcal{D}$. Because of the special relationship between the two distributions, we are asking whether it is possible for the network learned using data drawn from $\mathcal{D}_{\eta}$ to do well on data drawn from $\mathcal{D}$. As a matter of fact, we want it to do well on data only from $\mathcal{D}$; we do not want it to learn distribution $\mathcal{D}_{\eta}$. 
	
	Let $h$ and $h_{\eta}$ denote the classifier function (network) learned by an algorithm when given $S$ and $S_{\eta}$ as training data, respectively. 
	We can say that an algorithm \textbf{resists  memorization} if 
	\[
	\mbox{Prob}_{(X,y) \sim \mathcal{D}}[h(X) = y] =  \mbox{Prob}_{(X,y) \sim \mathcal{D}}[h_{\eta}(X) = y]
	\]
	What this means is that the accuracy on the original data achieved by the network learnt with noisy data is same as that of network learnt with original clean data. This is the ideal case where random altering of labels would have no effect on the classifier learnt. Note that the RHS above is what we called $J_2$ accuracy in the previous section. 
	
	The standard algorithm employed for training all networks is empirical risk minimization. The above property can be established for risk minimization if the loss function satisfies a special property called symmetry~\cite{ghosh2017aaai}.
	
	{\bf Definition}: A loss function $L$ is called {\bf symmetric} if it satisfies
	\[ \sum_{j=1}^K L(g(X), j) = C, \;\; \forall g, X\]
	where $C$ is a finite constant. That is, given any network (or function) $g$ and any input $X$, if we sum the loss values over all class labels, it should give the same constant. 
	
	\textbf{Theorem 1} Let $\mathcal{L}$ be a symmetric loss, $\mathcal{D}$ and $\mathcal{D}^\eta$ be as defined above. Assume $\eta < \frac{K-1}{K}$. Let $y_x$ and $\tilde{y}_x$ denote the original and noisy label corresponding to a pattern $X$.
	The risk of $h$ over $\mathcal{D}$ and over $\mathcal{D}^\eta$ is 
	$R_\mathcal{L}(h)  = \mathbb{E}_{\mathcal{D}}[\mathcal{L}(h(X), y_{x})]$ 
	and 
	$R_\mathcal{L}^\eta(h) = \mathbb{E}_{\mathcal{D}^\eta}[\mathcal{L}(h(X),  \tilde{y}_{x})]$ 
	respectively. Then, 
	given any two classifiers $h_1$ and $h_2$, if $R_\mathcal{L}(h_1) < R_\mathcal{L}(h_2)$, then $R_\mathcal{L}^\eta(h_1) < R_\mathcal{L}^\eta(h_2)$ and vice versa.
	
	\textbf{Proof} (This follows easily from the proof of Theorem~1 in \cite{ghosh2017aaai}.) 
	Given the way the randomized labels are generated, we have
	\begin{eqnarray*}
		R_{\mathcal{L}}^\eta(h)& = & \E_{{X}, \tilde{y}_{x}} \mathcal{L}(h(X), \Tilde{y}_{x}) \\
		&= & \E_{{X}} \E_{y_{x} | X} \E_{\Tilde{y}_{x} | X, y_{x}} \mathcal{L}(h(X), \Tilde{y}_{x}) \\
		& & \hspace{-1.5cm} = \E_{X} \E_{y_{x} | X} \left[ (1-\eta) \mathcal{L}(h(X),y_{x}) + \frac{\eta}{K-1} \sum_{i\neq y_{x}} \mathcal{L}(h(X),i) \right] \\
		& = & (1 - \eta) R_\mathcal{L}(h) +  \frac{\eta}{K-1} (C - R_\mathcal{L}(h))\\ 
		&=&  \frac{C \eta}{K-1} + \left(1-\frac{\eta K}{K-1}\right) R_\mathcal{L}(h) 	
	\end{eqnarray*}
	where $C$ is the constant in the symmetry condition on the loss function and K is the number of classes. Since $\eta < \frac{K-1}{K}$, we have $(1- \frac{\eta{K}}{K-1}) > 0$. Hence, the above shows that whenever $R_\mathcal{L}(h_1) < R_\mathcal{L}(h_2)$, we get $R^{\eta}_\mathcal{L}(h_1) < R^{\eta}_\mathcal{L}(h_2)$ and vice versa. This completes the proof.
	
	Theorem 1 shows that the symmetric loss maintains the risk ranking of different  networks regardless of random flipping of labels (as long as $\eta < \frac{K-1}{K}$). This implies that any local minimum of risk under randomly flipped labels would also be a local minimum of risk under original labels if the loss function is symmetric.

	
	
	The loss function RLL satisfies the symmetry condition \cite{rll}. Thus, if we are using RLL, then any local minimum of risk under $\mathcal{D}_{\eta}$ would also be a local minimum of risk under $\mathcal{D}$. Even though this result is only for minima of risk,  one can expect local minima of empirical risk under random label flips to be good approximators of local minima of empirical risk with clean, original samples. This explains the empirical results presented in the previous section  regarding the ability of RLL to resist memorization.  
	
	There are other losses that satisfy the symmetry condition, e.g., 0--1 loss, mean absolute value of error (MAE), etc. 
	
	It is easy to verify that neither CCE nor MSE satisfy the symmetry condition. Though the symmetry of loss is only a sufficient condition for robustness, this may provide an explanation of the overfitting observed with these loss functions when the labels are randomly flipped. 
	
	As is easy to see, the symmetry condition implies that the loss function is bounded. Given a bounded loss function we can satisfy the symmetry condition by `normalizing' it. Given a bounded loss $L$, define $\bar{L}$ by
	\[\bar{L}(g(X), j) = \frac{L(g(X), j)}{\sum_s L(g(X), s)}\]
	It is easy to see that $\bar{L}$ satisfies the symmetry condition. As mentioned earlier, CCE loss is unbounded and hence normalization would not turn it into a symmetric loss. However, we can normalize MSE loss. 
	
	In Figure \ref{fig:cifar-mnist-norm-mse} we show results obtained using normalized MSE. Once we normalize MSE, it no longer fits the data with random labels perfectly; the training accuracy now saturates at a value below $100\%$ and thus it behaves more like RLL now. 
	
	The empirical results presented in the previous section adequately demonstrate that the loss function can play a crucial role in mitigating the tendency of deep networks to memorize the training examples. The analysis presented here provides an explanation for this ability of RLL to resist such memorization. As a mater of fact, if the loss function is symmetric it would have such robustness and we can normalize a bounded loss to have such robustness.
	
	\section{Conclusions}
	Many recent studies have shown that overparameterized deep networks seem to be capable of perfectly fitting even randomly-labelled data. This phenomenon of memorization in deep networks has received a lot of attention because it raises important questions on how to understand generalization abilities of deep networks. In this paper we have shown through empirical studies that changing the loss function alone can significantly change the memorization in such deep networks. We showed this with the symmetric loss functions and we have provided some theoretical analysis to explain the empirical results. The results presented here suggest that choice of loss function can play a critical role in overfitting by deep networks.  We feel it is important to further investigate the nature of different loss functions for a better understanding of generalization abilities of deep networks. 
	\label{section:4}
	


	%
	%
	\bibliographystyle{splncs04}
	\bibliography{ref}
	
\end{document}